\documentclass[usenames,dvipsnames,sigconf]{acmart}

\usepackage[nolist]{acronym}
\AtBeginDocument{%
 \providecommand\BibTeX{{%
    \normalfont B\kern-0.5em{\scshape i\kern-0.25em b}\kern-0.8em\TeX}}}

\usepackage{soul}
\usepackage{color}
\usepackage{wallpaper}
\usepackage{graphicx}
\usepackage{cleveref}
\usepackage{numprint}
\usepackage{paralist}
\usepackage{xspace}
\usepackage[lofdepth,lotdepth]{subfig}

\definecolor{orangeX}{rgb}{1,.5,0}
\definecolor{blueX}{rgb}{.2, .59, .88}
\definecolor{purpleX}{rgb}{.294118, 0, .509804}
\definecolor{greenX}{rgb}{.421, .578, .241}
\definecolor{bole}{rgb}{0.47, 0.27, 0.23}
\definecolor{mypink3}{cmyk}{0, 0.7808, 0.4429, 0.1412}
\definecolor{mygray}{gray}{0.6}

\newcommand{\hume}{\textsc{Hume-Reaction}\,}

\newcommand{\musephysio}{\textsc{MuSe-Physio\,}}

\newcommand{\ds}{\textsc{DeepSpectrum}}
\newcommand{\opensmile}{\textsc{openSMILE}}

\newcommand{\pyfeat}{\textsc{Py-Feat\,}}

\newcommand{\vgg}{\textsc{VGGish}}
\newcommand{\vggf}{\textsc{VGGface 2}}

\newcommand{\mtcnn}{\textsc{MTCNN\,}}
\newcommand{\bert}{\textsc{BERT\,}}

\newcommand{\eg}{e.\,g., }

\newcommand{\cf}{{cf.\ }}

\copyrightyear{2022}
\acmYear{2022}
\setcopyright{acmcopyright}\acmConference[MuSe' 22]{Proceedings of the 3rd International Multimodal Sentiment Analysis Workshop and Challenge}{October 10, 2022}{Lisboa, Portugal}
\acmBooktitle{Proceedings of the 3rd International Multimodal Sentiment Analysis Workshop and Challenge (MuSe' 22), October 10, 2022, Lisboa, Portugal}
\acmPrice{15.00}
\acmDOI{10.1145/3551876.3554817}
\acmISBN{978-1-4503-9484-0/22/10}

\begin{document}
\title[MuSe 2022: Baseline Paper]{
The MuSe 2022 Multimodal Sentiment Analysis Challenge: Humor, Emotional Reactions, and Stress} %

\author{Lukas Christ}
\affiliation{%
  \institution{University of Augsburg}
  \city{Augsburg, Germany}}
  
\author{Shahin Amiriparian}
\affiliation{%
  \institution{University of Augsburg}
  \city{Augsburg, Germany}}

\author{Alice Baird}
\affiliation{%
  \institution{Hume AI}
  \city{New York, USA}}

\author{Panagiotis Tzirakis}
\affiliation{%
  \institution{Hume AI}
  \city{New York, USA}}

\author{Alexander Kathan}
\affiliation{%
  \institution{University of Augsburg}
  \city{Augsburg, Germany}}
  
  \author{Niklas Müller}
\affiliation{%
  \institution{University of Passau}
  \city{Passau, Germany}}
  
  \author{Lukas Stappen}
\affiliation{%
  \institution{Recoro}
  \city{Munich, Germany}}

\author{Eva-Maria Meßner}
\affiliation{%
  \institution{University of Ulm}
  \city{Ulm, Germany}}
  
\author{Andreas König}
\affiliation{%
  \institution{University of Passau}
  \city{Passau, Germany}}
  
\author{Alan Cowen}
\affiliation{%
  \institution{Hume AI}
  \city{New York, USA}}
  
\author{Erik Cambria}
\affiliation{%
  \institution{Nanyang Technological University}
  \city{Singapore}}

\author{Bj\"orn W. Schuller}
\affiliation{%
  \institution{Imperial College London}
  \city{London, United Kingdom}}
\renewcommand{\shortauthors}{Lukas Christ et al.}

\settopmatter{printacmref=true}

\begin{abstract}

The \ac{MuSe} 2022 is dedicated to multimodal sentiment and emotion recognition. For this year's challenge, we feature three datasets: (i) the \ac{Passau-SFCH} dataset that contains audio-visual recordings of German football coaches, labelled for the presence of humour; (ii) the \hume dataset in which reactions of individuals to emotional stimuli have been annotated with respect to seven emotional expression intensities, and (iii) the \ac{Ulm-TSST} dataset comprising of audio-visual data labelled with continuous emotion values (arousal and valence) of people in stressful dispositions. Using the introduced datasets, \ac{MuSe} 2022 addresses three contemporary affective computing problems: in the \ac{MuSe-Humor}, spontaneous humour has to be recognised; in the \ac{MuSe-Reaction}, seven fine-grained `in-the-wild' emotions have to be predicted; and in the \ac{MuSe-Stress}, a continuous prediction of stressed emotion values is featured. 
The challenge is designed to attract different research communities, encouraging a fusion of their disciplines. Mainly, \ac{MuSe} 2022 targets the communities of audio-visual emotion recognition, health informatics, and symbolic sentiment analysis. This baseline paper describes the datasets as well as the feature sets extracted from them. A recurrent neural network with \acs{LSTM} cells is used to set competitive baseline results on the test partitions for each sub-challenge. We report an \ac{AUC} of $.8480$ for \ac{MuSe-Humor}; $.2801$ mean (from 7-classes) Pearson's Correlations Coefficient ($\rho$) for \ac{MuSe-Reaction}, as well as $.4931$ \ac{CCC} and $.4761$ for valence and arousal in \ac{MuSe-Stress}, respectively.

\end{abstract}

\ccsdesc[500]{Computing methodologies~Neural networks}
\ccsdesc[500]{Computing methodologies~Artificial intelligence}
\ccsdesc[300]{Computing methodologies~Computer vision}
\ccsdesc[300]{Computing methodologies~Natural language processing}

\keywords{Multimodal Sentiment Analysis; Affective Computing; Humor Detection; Emotion Recognition; Multimodal Fusion; Challenge; Benchmark}

\maketitle
\section{Introduction}
The 3rd edition of the \textbf{Mu}ltimodal \textbf{Se}ntiment Analysis (MuSe) Challenge addresses three tasks:  
humour detection and categorical as well as dimensional emotion recognition. Each corresponding sub-challenge utilises a different dataset. In the 
\acl{MuSe-Humor} (\textbf{\ac{MuSe-Humor}}), participants will detect the presence of humour in football press conference recordings. For \ac{MuSe-Humor}, the novel \acl{Passau-SFCH} (\textbf{\ac{Passau-SFCH}}) dataset is introduced. It features press conference recordings of 10 German Bundesliga football coaches, recorded between August 2017 and November 2017. Initially, the dataset comprises about 18 hours of video, where each of the 10 coaches accounts for at least 90 minutes of data. The subset provided in the challenge  still features 11 hours of video. Originally, the data is annotated for direction as well as sentiment of humour following the two-dimensional model of humour proposed in \cite{martin2003individual}. In the challenge, only the presence of humour is to be predicted.

For the Emotional Reactions Sub-Challenge (\ac{MuSe-Reaction}), emotional reactions are explored by introducing a first of its kind, large-scale (2,222 subjects, 70+ hours), multi-modal (audio and video) dataset: \hume. The data was gathered in the wild, with subjects recording their own facial and vocal reactions to a wide range of emotionally evocative videos via their webcam, in a wide variety of at-home recording settings with varying noise conditions. Subjects selected the emotions they experienced in response to each video out of 48 provided categories and rated each selected emotion on a 0-100 intensity scale. In this sub-challenge, participants will apply a multi-output regression to predict the intensities of seven self-reported emotions from the subjects' multi-modal recorded responses: \begin{inparaitem}[]\item Adoration, \item Amusement, \item Anxiety, \item Disgust, \item Empathic Pain, \item Fear, \item Surprise \end{inparaitem}. 

The \acl{MuSe-Stress} (\textbf{\ac{MuSe-Stress}}) is a regression task on continuous signals for valence and arousal. It is based on the \acl{Ulm-TSST} dataset (\ac{Ulm-TSST}), comprising individuals in a stress-inducing scenario following the \ac{TSST}. This sub-challenge is motivated by the prevalence of stress and its harmful impacts in modern societies~\cite{can2019stress}. In addition to audio, video and textual features, \ac{Ulm-TSST} includes four biological signals captured at a sampling rate of 1\,kHz; EDA, Electrocardiogram (ECG), Respiration (RESP), and heart rate (BPM). \ac{MuSe-Stress} was already part of MuSe 2021~\cite{stappen2021muse}, where it attracted considerable interest. Due to some participants reporting challenges
generalising to the test set~\cite{hamieh2021multi, duong2021multi}, we rerun the challenge, allowing participants to submit more predictions than in the previous iteration. We thereby hope to encourage participants to thoroughly explore the robustness of their proposed approaches. Moreover, for this year's \ac{MuSe-Stress} sub-challenge, we use the labels of last year's \musephysio sub-challenge as the arousal gold standard.

\begin{table}[t!]
\footnotesize
  \caption{ Reported are the number (\#) of unique subjects, and the duration for each sub-challenge hh\,:mm\,:ss. 
\label{tab:paritioning}
 }
 \resizebox{\linewidth}{!}{
  \begin{tabular}{lrcrcrc}
    \toprule
     & \multicolumn{2}{c}{\textbf{\ac{MuSe-Humor}}} & \multicolumn{2}{c}{\textbf{\ac{MuSe-Reaction}}} & \multicolumn{2}{c}{\textbf{\ac{MuSe-Stress}}}\\
     \cmidrule(lr){2-3} \cmidrule(lr){4-5} \cmidrule(lr ){6-7}
    Partition & \# & Duration & \# &  Duration & \# & Duration \\
    \midrule
    Train   & 4 & 3\,:52\,:44 & 1334 &51\,:04\,:02 & 41 & 3\,:25\,:56 \\
    Development  &  3 & 3\,:08\,:12 & 444 & 14\,:59\,:27 & 14 & 1\,:10\,:50  \\
    Test    &  3 & 3\,:55\,:41 & 444 & 14\,:48\,:21 & 14 & 1\,:10\,:41 \\
    \midrule
    $\sum$    & 10 & 10\,:56\,:37 & 2222 & 74\,:26\,:19 & 69 & 5\,:47\,:27 \\
  \bottomrule
\end{tabular}
}
\end{table}

By providing the mentioned tasks in the 2022 edition of \ac{MuSe}, we aim for addressing research questions that are of interest to affective computing, machine learning and multimodal signal processing communities and encourage a fusion of their disciplines. Further, we hope that our multimodal challenge can yield new insights into the merits of each of the core modalities, as well as various multimodal fusion approaches. Participants are allowed to use the provided feature sets in the challenge packages and integrate them into their own machine learning frameworks.

The paper is structured as follows: \Cref{sec:challenges} introduces the three sub-challenges alongside with the datasets they are based on, and outlines the challenge protocol. Then, pre-processing, provided features, their alignment, and our baseline models are described in \Cref{sec:features}. In \Cref{sec:results}, we present and discuss our baseline results before concluding the paper in \Cref{sec:conclusion}.

A summary of the challenge results can be found in \cite{Amiriparian22-TM2}.

\section{The Three Sub-Challenges}\label{sec:challenges}
In what follows, each sub-challenge and dataset is described in detail, as well as the participation guidelines.

\begin{figure}[h!]
    \centering
    \subfloat[Valence label distribution]{
    \includegraphics[width=.45\columnwidth]{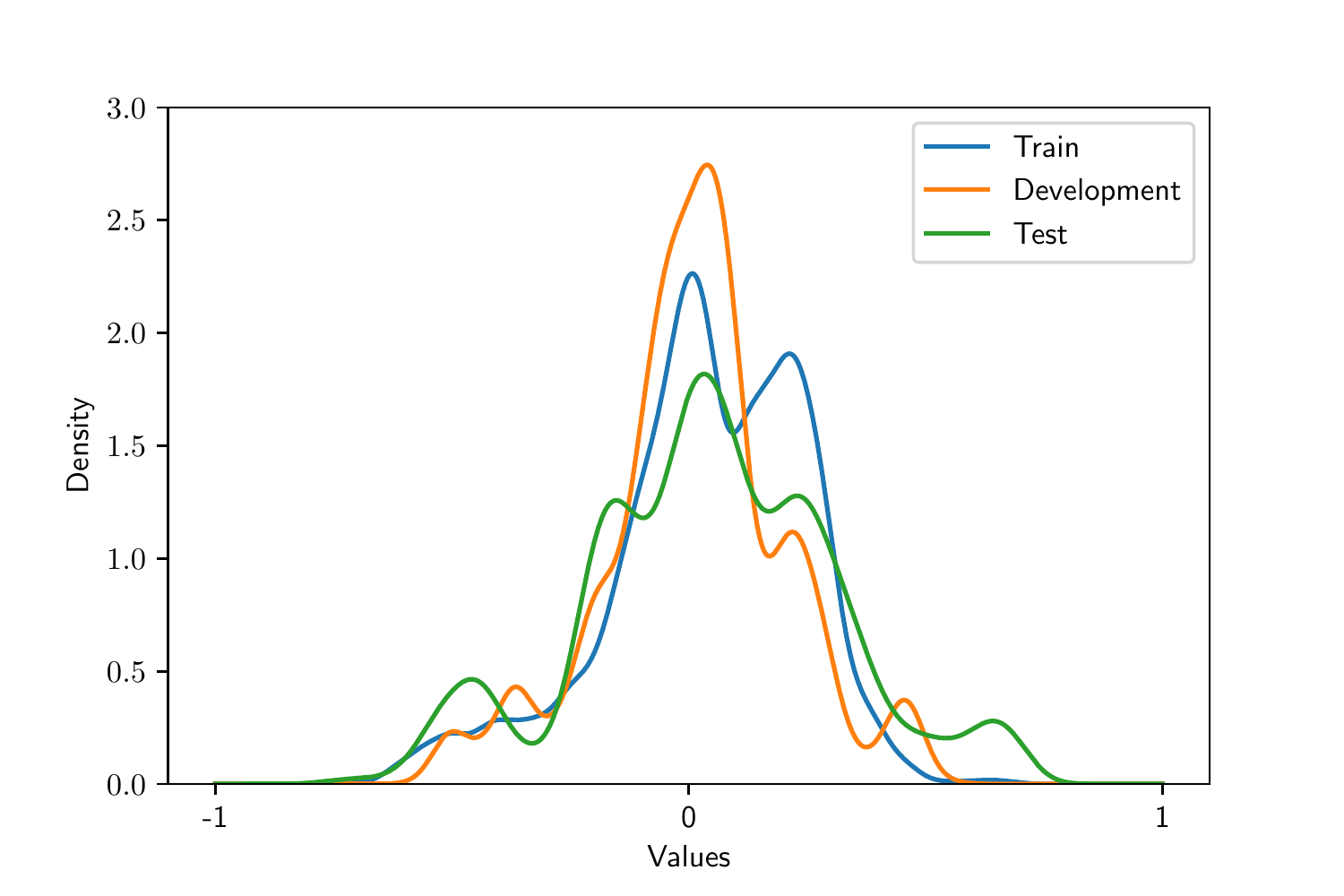}}
    \label{fig:density-valence}
    \subfloat[Physiological arousal label distribution]{
    \includegraphics[width=.45\columnwidth]{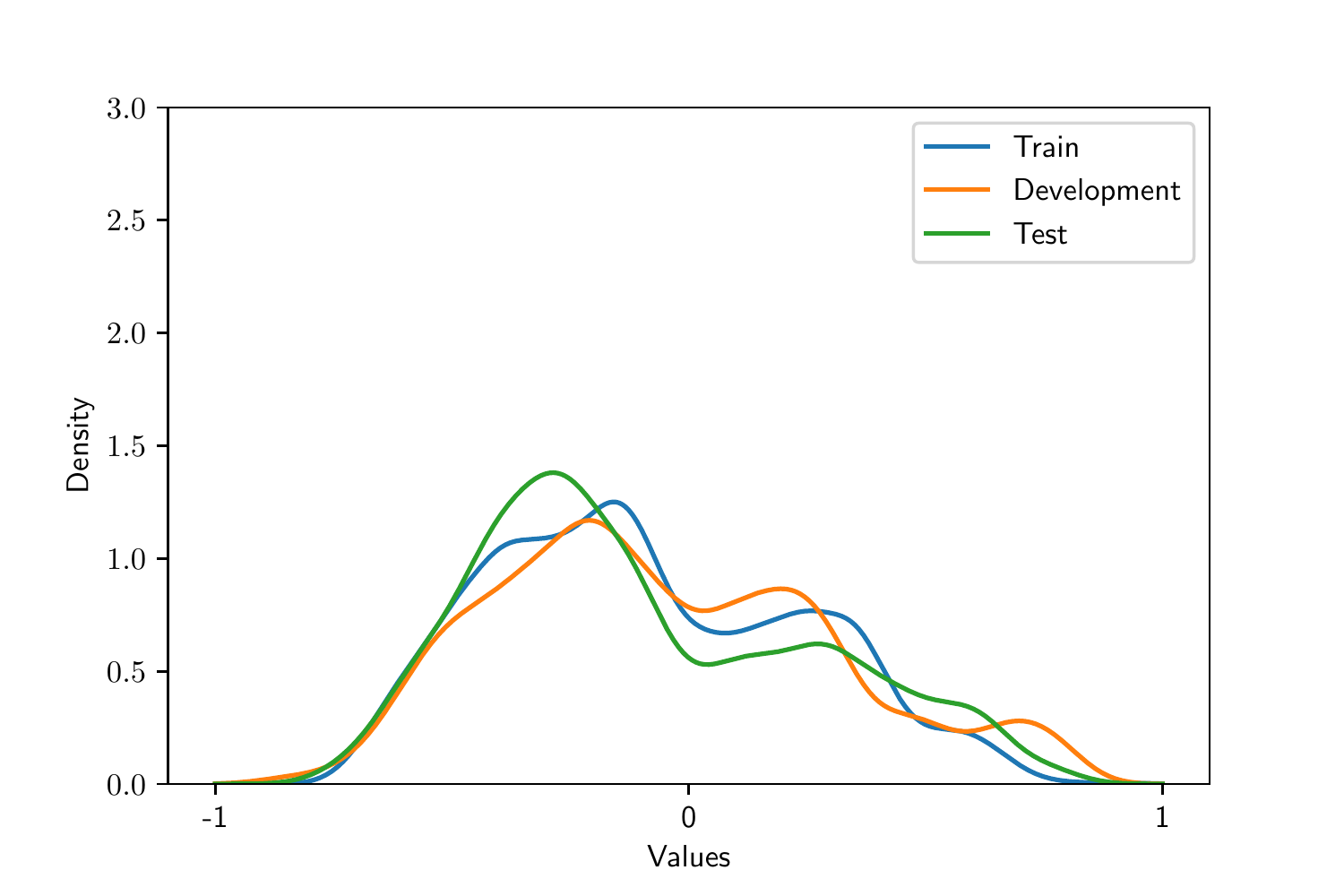}}
    \label{fig:density-arousal}
    \caption{Frequency distribution in the partitions train, development, and test for the continuous prediction sub-challenge \ac{MuSe-Stress}.}
    \label{fig:freq}
\end{figure}

\subsection{The MuSe-Humor Sub-Challenge\label{sec:humor}} 
Humour is one of the richest and most consequential elements of
human behaviour and cognition~\cite{gkorezis2014leader} and thus of high relevance in the context of affective computing and human-computer interaction. As humour can be expressed both verbally and non-verbally, multimodal approaches are especially suited for detecting humour. However, while humour detection is a very active field of research in Natural Language Processing (\eg \cite{chen2018humor, yang2015humor}), only a few multimodal datasets for humour detection exist~\cite{hasan2019ur, mittal2021so, wu2021mumor}. Especially, to the best of our knowledge, there are no datasets for detecting humour in spontaneous, non-staged situations. With \ac{MuSe-Humor}, we intend to address this research gap.

In this challenge, the \ac{Passau-SFCH} dataset is utilised. It features video and audio recordings of press conferences of 10 German Bundesliga football coaches, during which the coaches occasionally express humour. The press conferences present natural, live, semi-staged communication of the coaches to and with journalists in the audience. All subjects are male and aged between 30 and 53 years. The dataset is split into speaker independent partitions. The training set includes the videos of 4 coaches, while the development and test partition both comprise the videos of 3 coaches. We only include segments in which the coach is speaking to ensure that humour is detected from the behaviour of the coach and not the audience, \eg laughter. Participants are provided with video, audio, and ASR transcripts of said segments. To obtain the transcripts, we utilise a Wav2Vec2-Large-XLSR~\cite{conneau2020unsupervised} model fine-tuned on the German data in Common Voice~\cite{commonvoice:2020}\footnote{\href{https://huggingface.co/jonatasgrosman/wav2vec2-large-xlsr-53-german}{https://huggingface.co/jonatasgrosman/wav2vec2-large-xlsr-53-german}}. Moreover, manually corrected transcripts are included.

Every video was originally labelled by 9 annotators at a 2\,Hz rate indicating sentiment and direction of the humour expressed, as defined by the two-dimensional humour model proposed by~\citet{martin2003individual} in the \ac{HSQ}. For the challenge, we only build upon binary humour labels, i.\,e., indicating if the coach's communication is humorous or not. We obtain a binary label referring to presence or absence of humour using the following three steps. First, we only consider the humour dimension label of sentiment. Second, based on the sentiment labels, we filter out annotators displaying low agreement with other annotators. In order to account for slight lags in annotation signals, we choose to compute the target humour labels for frames of two seconds using a step size of one second. Finally, such a frame is considered as containing humour if at least 3 of the remaining annotators indicate humour within this frame. As a result, 4.9\,\% of the training partition frames, 2.9~\% of the development partition frames, and 3.9\,\% of the test partition frames are labelled as humorous. We deliberately opted for a split in which the humour label is over-represented in the training partition in order to help participants' models with learning. The provided features are extracted at 2\,Hz rates. They can easily be mapped to the 2\,s segments they belong to.

For evaluation, the \ac{AUC} metric is utilised, indicating how well a model can separate humorous from non-humorous frames.

\subsection{The MuSe-Reaction Sub-Challenge\label{sec:reaction}}

Computational approaches for understanding human emotional reactions are of growing interest to researchers~\cite{kumar2021construction,sun2020eev}, with emerging applications ranging from pedagogy~\cite{chalfoun2006predicting} to medicine~\cite{skoraczynski2017predicting}. A person's reaction to a given stimulus can be informative about both the stimulus itself, \eg whether educational material is interesting to a given audience, and about the person, \eg their level of empathy~\cite{tamborini1990reacting} and well-being~\cite{zohar2005effects}. However, progress in developing computational approaches to understand human emotional reactions has been hampered by the limited availability of large-scale datasets of spontaneous emotional reactions. Thus, for the \ac{MuSe-Reaction} sub-challenge, we introduce the \ac{Hume-Reaction} dataset, which consists of more than 70 hours of audio and video data, from 2,222 subjects from the United States (1,138) and South Africa (1,084), aged from 18.5 -- 49.0 years old.

The subjects within the dataset are reacting to a wide range of emotionally evocative stimuli (2,185 stimuli in total~\cite{cowen2017self}). Each sample within the dataset has been self-annotated by the subjects themselves for the intensity of 7 emotional expressions in a range from 1-100: \begin{inparaitem}[]\item Adoration, \item Amusement, \item Anxiety, \item Disgust, \item Empathic Pain, \item Fear, \item Surprise\end{inparaitem}. 

The data is self-recorded via subjects' own webcams in an environment of their choosing, including a wide variety of background, noise, and lighting conditions. Furthermore, different subjects spontaneously reacted with their faces and voices to varying degrees, such that the audio and multi-modal aspects of this sub-challenge will be particularly interesting to incorporate. The organisers also provide labels for detected (energy-based) vocalisations to aid participants in incorporating audio, with a total of 8,064 multi-modal recordings found to include vocalisations.

For the \ac{MuSe-Reaction} sub-challenge the aim is to perform a multi-output regression from features extracted from the multi-modal (audio and video) data for the intensity of 7 emotional reaction classes. For this sub-challenge's evaluation, the Pearson's correlations coefficient  ($\rho$) is reported as the primary baseline.

\subsection{The \ac{MuSe-Stress} Sub-challenge}

The \ac{MuSe-Stress} task is based on the multimodal \ac{Ulm-TSST} database, for which subjects were recorded in a stress-inducing, free speech scenario, following the \ac{TSST} protocol~\cite{kirschbaum1993trier}. In the \ac{TSST}, a job interview situation is simulated. Following a short period of preparation, a five-minute free speech oral presentation is given by the subjects. This presentation is supervised by two interviewers, who do not communicate with the subjects during the five minutes. 
\ac{Ulm-TSST} comprises recordings of such \ac{TSST} presentations of 69 participants (49 of them female), aged between 18 and 39 years. Overall, \ac{Ulm-TSST} includes about 6 hours of data (\cf \Cref{tab:paritioning}). On the one hand, the dataset features the audio, video, and text modalities. On the other hand, the physiological signals ECG, RESP, and BPM are provided. For extensive experiments on multimodal emotion recognition in \ac{TSST}-based multimodal datasets see~\cite{alice_tsst}.

\ac{Ulm-TSST} has been annotated by three raters continuously for the emotional dimensions of valence and arousal, at a 2\,Hz sampling rate. Regarding valence, a gold standard is created by fusing the three corresponding annotator ratings, utilising the \ac{RAAW} method from the MuSe-Toolbox~\cite{stappen2021toolbox}. \ac{RAAW} addresses the difficulties arising when emotion annotations -- subjective in their nature -- are to be combined into a gold standard signal. In short, \ac{RAAW} first tackles the inherent rater lag by aligning the (per annotator) standardised signals via generalised \ac{CTW}~\cite{zhou2015generalized}. After that, the \ac{EWE}~\cite{grimm2005evaluation} is applied to the aligned signals. \ac{EWE} fuses the individual signals using a weighting based on each rater's inter-rater agreement to the mean of all others. A detailed description of \ac{RAAW} can be found in \cite{stappen2021toolbox}. We obtain a mean inter-rater agreement of 0.204 ($\pm$ 0.200) for valence.

As for the arousal gold standard, a different approach is employed. Instead of fusing the three annotators' arousal ratings, we take the labels of last year's \musephysio sub-challenge as the arousal gold standard. Here, the annotator with lowest inter-rater agreement is discarded and replaced with the subject's electrodermal activity signal (EDA) which is known to indicate emotional arousal~\cite{caruelle2019use}. This signal is downsampled to 2\,Hz and smoothed using a Savitzky–Golay filtering approach (window size of 26 steps) in advance. Then, the two remaining annotators and the preprocessed EDA signal are again fused via \ac{RAAW}, resulting in a mean inter annotator agreement of 0.233 ($\pm 0.289$). This signal is called \emph{physiological arousal} in the following. The motivation to employ this kind of gold standard is to obtain a more objective arousal signal. Considering such an objective criterion for arousal in addition to subjective annotations is especially relevant given the task at hand: in the job interview setting, individuals can be expected to try to hide their arousal, making it more difficult for annotators to recognise it. Detailed experiments on combining subjective annotations with objective physiological signals are provided in~\cite{baird2021physiologically}.

\ac{Ulm-TSST} is split into train, development, and test partitions containing 41, 14, and 14 videos, respectively. The split is identical to the split used in last year's challenge. \Cref{fig:freq} shows the distributions of the valence and physiological arousal signals for the dataset.

\subsection{Challenge Protocol}
All challenge participants are required to complete the \ac{EULA} which is available on the \ac{MuSe} 2022 homepage\footnote{\href{https://www.muse-challenge.org}{https://www.muse-challenge.org}}. Further, the participants must hold an academic affiliation. Each challenge contribution should be followed by a paper 
that describes the applied methods and provides the obtained results. The peer review process is double-blind. To obtain results on the test set, participants upload their predictions for unknown test labels on CodaLab\footnote{ \href{https://codalab.lisn.upsaclay.fr/}{https://codalab.lisn.upsaclay.fr/}}.
The number of prediction uploads depends on the sub-challenge: for \ac{MuSe-Humor} and \ac{MuSe-Reaction}, up to 5 prediction uploads can be submitted, while for \ac{MuSe-Stress}, up to 20 prediction uploads are allowed. We want to stress that the organisers themselves do not participate as competitors in the challenge.

\section{Baseline Features and Model}\label{sec:features}
To enable the participants to get started quickly, we provide a set of features extracted from each sub-challenge's video data. More precisely, the provided features include of up to five model-ready video, audio, and linguistic feature sets, depending on the sub-challenge\footnote{Note: Participants are free to use other external resources such as features, datasets, or pretrained networks. The accompanying paper is expected to clearly state and explain the sources and tools used.}. 
Regarding the label sampling rate, labels refer to 2\,s windows in \ac{MuSe-Humor}. The \ac{MuSe-Stress} data is labelled at a 2\,Hz rate. For \ac{MuSe-Reaction}, there is one label vector of 7 classes per sample.

\subsection{Pre-processing}
All datasets are split into training, development, and test sets. For all partitions, ratings, speaker independence, and duration are taken into consideration (\cf \Cref{tab:paritioning}). The videos in \ac{Passau-SFCH} are cut to only include segments in which the respective coach is actually speaking. As the press conference setting can be seen as a dialogue between journalists and the coach, the answers given by each coach provide a natural segmentation of the \ac{Passau-SFCH} data. 
For \ac{MuSe-Reaction} -- as can be seen in \Cref{tab:paritioning} --, a 60-20-20\% split strategy is applied. There is no additional segmentation applied to clean the data further, each sample contains a single reaction to an emotional stimulus, and labels were normalised per sample to range from [0\,:1]. For further exploration, the participants are also provided with voice activity segments from the samples, which show to contain audio of substantial energy.
In the \ac{Ulm-TSST} dataset, we make sure to exclude scenes which are not a part of the TSST setting, \eg the instructor speaking. Moreover, we cut segments in which TSST participants reveal their names. The \ac{Ulm-TSST} dataset is not segmented any further.

\subsection{Audio}
All audio files are first normalised to -3 decibels and then converted from stereo to mono, at 16\,kHz, 16\,bit. Afterwards, we make use of the two well-established machine learning toolkits \opensmile{}~\cite{eyben2010opensmile} and \ds{}~\cite{Amiriparian17-SSC} for expert-designed and deep feature extraction from the audio recordings. Both systems have proved valuable in audio-based \ac{SER} tasks~\cite{Amiriparian22-DAP,Gerczuk22-EAT,Schuller21-TI2}.

\subsubsection{\acs{eGeMAPS}}
\label{ssec:egemaps}
The \opensmile{} toolkit~\cite{eyben2010opensmile}\footnote{\href{https://github.com/audeering/opensmile}{https://github.com/audeering/opensmile}} is used for the extraction of the \ac{eGeMAPS}~\cite{eyben2015geneva}. This feature set which is proven valuable for \ac{SER} tasks~\cite{baird2019can}, also in past MuSe challenges (\eg \cite{vlasenko2021fusion}), includes 88 acoustic features that can capture affective physiological changes in voice production.  In \ac{MuSe-Humor}, we use the default configuration to extract the 88 \ac{eGeMAPS} functionals for each two second audio frame. For the audio of \ac{MuSe-Reaction}, the 88 \ac{eGeMAPS} functionals are extracted with a step size of 100\,ms and window size of 1 second. Regarding \ac{MuSe-Stress}, the functionals are obtained with a 2\,Hz rate, using a window size of 5 seconds.

\subsubsection{\ds}

The principle of \ds~\cite{Amiriparian17-SSC}\footnote{\href{https://github.com/DeepSpectrum/DeepSpectrum}{https://github.com/DeepSpectrum/DeepSpectrum}} is to utilise pre-trained image \acp{CNN} for the extraction of deep features from visual representations (\eg Mel-spectrograms) of audio signals. The efficacy of \ds{} features has been demonstrated for \ac{SER}~\cite{Ottl20-GSE}, sentiment analysis~\cite{Amiriparian17-SAU}, and general audio processing tasks~\cite{Amiriparian20-TCP}. For our \ds{} baseline experiments, we use \textsc{DenseNet121}~\cite{huang2017densely} pre-trained on ImageNet~\cite{russakovsky2015imagenet} as the \ac{CNN} backbone. The audio is represented as a Mel-spectrogram with $128$ bands employing the viridis colour mapping. Subsequently, the spectrogram representation is fed into \textsc{DenseNet121}, and the output of the last pooling layer is taken as a $1\,024$-dimensional feature vector. The window size is set to one second, the hop-size to $500$\,ms.

\subsection{Video}
To extract specific image descriptors related to facial expressions, we make use of two \ac{CNN} architectures: \ac{MTCNN} and \vggf{}. We also provide a set of \acp{FAU} obtained from faces of individuals in the datasets. Further, participants are also given the set of extracted faces from the raw frames. In the videos of \ac{MuSe-Humor}, typically more than one face is visible. As this sub-challenge's objective is to predict the expression of humour of the coach, we only provide the faces of the respective coach and the features computed for them.

\subsubsection{\acs{MTCNN}}
The \ac{MTCNN}~\cite{zhang2016mtcnn} model\footnote{\url{https://github.com/ipazc/mtcnn}}, pre-trained on the data\-sets WIDER FACE~\cite{yang2016wider} and CelebA~\cite{liu2015faceattributes}, is used to detect faces in the videos. 
Two steps are carried out to filter extracted faces that do not show the coach in \ac{Passau-SFCH}: first, we automatically detect the respective coach's faces using FaceNet\footnote{\href{https://github.com/timesler/facenet-pytorch}{https://github.com/timesler/facenet-pytorch}} embeddings of reference pictures showing the coach. The results of this procedure are then corrected manually. \ac{Ulm-TSST}, in contrast, has a simple, static setting. The camera position is fixed and videos only show the \ac{TSST} subjects who typically do not move much. Similarly, for \ac{MuSe-Reaction}, the video is captured from a fixed webcam. Hence, the performance of \mtcnn is almost flawless for both \ac{Hume-Reaction} and \ac{Ulm-TSST}. The extracted faces then serve as inputs of the feature extractors \vggf{} and \pyfeat.

\subsubsection{\vggf}
The purpose of \vggf{} is to compute general facial features for the previously extracted faces. \vggf{}~\cite{cao2018vggface2} is a dataset for the task of face recognition. It contains 3.3 million faces of about 9,000 different persons. As the dataset is originally intended for supervised facial recognition purposes, models trained on it compute face encodings not directly related to emotion and sentiment. We use a ResNet50~\cite{he2016deep} trained on \vggf{}\footnote{\href{https://github.com/WeidiXie/Keras-VGGFace2-ResNet50}{https://github.com/WeidiXie/Keras-VGGFace2-ResNet50}}  and detach its classification layer, resulting in a 512-dimensional feature vector output referred to as \vggf{} in the following.

\subsubsection{FAU}
\acp{FAU} as originally proposed by Ekman and Friesen~\cite{ekman1978facial}, are closely related to the expression of emotions. 
Hence, detecting \acp{FAU} is a promising and popular approach to the visual prediction of affect-related targets (\eg \cite{mallol2020investigation}). 
We employ \pyfeat\footnote{\url{https://py-feat.org}} to obtain predictions for the presence of 20 different \acp{FAU}. We do not change \pyfeat's default configuration, so that a pre-trained random forest model is used to predict the \acp{FAU}.

\subsection{Language: Bert}
 
In recent years, pre-trained Transformer language models account for state-of-the-art results in numerous Natural Language Processing tasks, also in tasks related to affect (\eg ~\cite{Schuller21-TI2}). In general, these models are pretrained in a self-supervised way utilising large amounts of text data. Subsequently, they can be fine-tuned for specific downstream tasks. For the transcripts of \ac{MuSe-Humor} and \ac{MuSe-Stress}, we employ a German version of the \bert (Bidirectional Encoder Representations from Transformers ~\cite{devlin2019bert}) model\footnote{\hyperlink{https://huggingface.co/bert-base-german-cased}{https://huggingface.co/bert-base-german-cased}}. No further fine-tuning is applied.  For both \ac{Passau-SFCH} and \ac{MuSe-Stress}, we extract the \bert token embeddings. Additionally, we obtain 768 dimensional sentence embeddings for all texts in \ac{Passau-SFCH} by using the encodings of \bert's $[CLS]$ token. In all cases, we average the embeddings provided by the last 4 layers of the \bert model, following ~\cite{sun2020multi}.

\subsection{Alignment}

For each task, at least two different modalities are available. Typically, sampling rates per modality may differ. We sample the visual features with a rate of 2\,Hz in all sub-challenges. The only exception is the \acp{FAU} in \ac{MuSe-Reaction}, which are sampled at a 4\,Hz rate. Regarding the audio features (\ds{} and \ac{eGeMAPS}{}), we apply the same frequency in \ac{MuSe-Humor} and \ac{MuSe-Stress}, while \ac{eGeMAPS} features are obtained using a step size of 100\,ms in \ac{MuSe-Reaction}. As \vgg{} and \acp{FAU} are only meaningful if the respective frame actually includes a face, we impute frames without a face with zeros.

For \ac{MuSe-Humor}, the binary humour label refers to frames of at most 2 seconds length. Hence, each label in \ac{MuSe-Humor} corresponds to at most 4 facial and acoustic feature vectors. 2\,Hz sentence embedding vectors are constructed by assigning every sentence to the 500\,ms frames it corresponds to. If two sentences fall into the same frame, their embeddings are averaged to form the feature for that frame.

Regarding \ac{MuSe-Reaction}, there is no alignment needed with labels, as each file is associated to a single vector of 7 emotional reaction labels.

For the \ac{MuSe-Stress} sub-challenge, we provide label-aligned features. Hence, these features exactly align with the labels. We apply zero-padding to the frames, where the feature type is absent. Moreover, we downsample the biosignals in \ac{Ulm-TSST} to 2\,Hz, followed by a smoothing utilising a Savitzky-Golay filter. Participants are provided with both the raw signals and the downsampled ones.

In both \ac{Ulm-TSST} and \ac{Passau-SFCH}, manual transcripts are available. However, they lack timestamps. Hence, we reconstruct word level timestamps utilising the Montreal Forced Aligner (MFA)~\cite{mcauliffe2017montreal} tool. Here, we employ the German (Prosodylab) model and the German Prosodylab dictionary.
The text features are then aligned to the 2\,Hz label signal by repeating each word embedding throughout the determined interval of the corresponding word. In case a 500\,ms frame comprises more than one word, we average over the word embeddings. Zero imputing is applied to parts where subjects do not speak. For the sentence embeddings in \ac{Passau-SFCH} we choose an analogous approach, repeating and, if applicable, averaging the embeddings.

\subsection{Baseline Model: LSTM-RNN\label{sec:model}}
The sequential nature of the tasks makes recurrent neural networks (RNNs) a natural choice for a fairly simple baseline system. More specifically, we employ a Long Short-Term Memory (LSTM)-RNN. Initially, we train a single model on each of the available feature sets. Regarding \ac{MuSe-Stress}, we separately train a model for both labels, valence and physiological arousal. We conduct an extensive hyperparameter search for each prediction target and feature. We thus optimise the number of RNN layers, the dimensionality of the LSTM's hidden vectors and the learning rate. Of note, we also experiment with both unidirectional and bidirectional LSTMs. The code as well as the configurations found in the hyperparameter search are available in the baseline GitHub repository\footnote{ \href{https://github.com/EIHW/MuSe2022}{https://github.com/EIHW/MuSe2022}}.

Each label in \ac{MuSe-Humor} is predicted based on all feature vectors belonging to the corresponding 2\,s window. Hence, the sequence length in the \ac{MuSe-Humor} training process is at most 4 steps.

In both \ac{MuSe-Reaction} and \ac{MuSe-Stress}, we make use of a segmentation approach which showed to improve results in previous works~\cite{sun2020multi, stappen2020muse1,stappen2021multimodal}. We find that a segmentation of the training data with a window size of 50\,s (i.\,e., 200 steps) and a hop size of 25\,s (i.e., 100 steps) leads to good results for \ac{MuSe-Stress}. For \ac{MuSe-Reaction} a slightly larger size of 500 steps and a hop size of 250, lead to more robust results. 

Following the unimodal experiments, in order to combine different modalities, for \ac{MuSe-Humor} and \ac{MuSe-Stress}, we implement a simple late fusion approach. We apply the exact same training procedure as before, now treating the predictions of previously trained unimodal models as input features. In these experiments, we use one configuration per task, without performing a hyperparameter search for every possible modality combination in \ac{MuSe-Stress}. As this approach for late fusion is less suited to a multi-label strategy, we apply an early fusion strategy for \ac{MuSe-Reaction}. For early fusion, we simply concatenate the best performing feature sets for each modality (audio and video), and then train a new model with the same hyperparameters from the uni-modal experiments. The code and configuration for the two fusion methods are also part of the baseline GitHub repository\footnote{\href{https://github.com/EIHW/MuSe2022}{https://github.com/EIHW/MuSe2022}}.
Moreover, the repository also includes links to the best model weight files in order to ease reproducibility.

\section{Experiments and Baseline Results}\label{sec:results}

We apply the model described above for every sub-challenge. In what follows, we discuss the baseline results in more detail.

\subsection{\ac{MuSe-Humor}}
The results for \ac{MuSe-Humor} are given in~\Cref{tab:humor}. Each result is obtained from running the LSTM using the specified features with 5 different fixed seeds, consistent with the challenge setting.

\begin{table}[h!]
\caption{Results for \ac{MuSe-Humor}. We report the AUC-Scores for the best among 5 fixed seeds, as well as the mean \ac{AUC}-Scores over these seeds and the corresponding standard deviations.}

\resizebox{1\columnwidth}{!}{%

\centering
 \begin{tabular}{lcc}
 \toprule 
 & \multicolumn{2}{c}{[\ac{AUC}]} \\
 Features & \multicolumn{1}{c}{Development} & \multicolumn{1}{c}{Test}  \\ \midrule \midrule
 \multicolumn{3}{l}{\textbf{Audio}} \\
 \ac{eGeMAPS} & .6861 (.6731 $\pm$ .0172) & .6952 (.6979 $\pm$ .0098) \\
 \ds & .7149 (.7100 $\pm$ .0030) & .6547 (.6497 $\pm$ .0102) \\
 \midrule
  \multicolumn{3}{l}{\textbf{Video}} \\
 \ac{FAU} & .9071 (.9030 $\pm$ .0028) & .7960 (.7952 $\pm$ .0077) \\
 \vggf & .9253 (.9225 $\pm$ .0024) & \textbf{.8480} (.8412 $\pm$ .0027) \\
 \midrule
 \multicolumn{3}{l}{\textbf{Text}} \\
 \textsc{BERT} & .8270 (.8216 $\pm$ 0045) & .7888 (.7905 $\pm$ 0035)
 \\
 \midrule
 \multicolumn{3}{l}{\textbf{Late Fusion}} \\
 A+T & .8901 (.8895 $\pm$ .0005) & .7804 (.7843 $\pm$ .0037) \\
 A+V & .8252 (.8219 $\pm$ .0038) & .6643 (.6633 $\pm$ .0027) \\
 T+V & .8908 (.8893 $\pm$ .0015) & .8232 (.8212 $\pm$ .0017) \\
 A+T+V & .9033 (.9026 $\pm$ .0006) & .7973 (.7910 $\pm$ .0057) \\
 \bottomrule
 \end{tabular}\label{tab:humor}
}

\end{table}

Evaluating audio and video features for the \ac{MuSe-Humor} sub-challenge shows a clear pattern. The video-based features, \ac{FAU} and \vgg{}, clearly outperform the audio-based features with \vgg{} accounting for an \ac{AUC} of $.8480$ on the test set while \ac{eGeMAPS} only achieves $.6952$ \ac{AUC}. This comes as no surprise, given that the expression of humour is often accompanied by smile or laughter and thus recognisable from facial expressions features. A manual inspection of the humorous segments confirms this intuition. Nevertheless, audio features are able to detect humour, too. Partly, this may be due to the presence of laughter. The performance of text features ($.7888$ on the test set) is slightly worse than for the features based on the video modality, but also better than the performance of the audio features. We find that the sentence-level \textsc{BERT} features outperform the token-level features. With the simple fusion of modalities, the performance is not improved. Specifically, the late fusion approach typically shows worse generalisation to the test data than the unimodal experiments. \eg there is a discrepancy of about $.16$ between mean \ac{AUC} on the development ($.8219$) and test ($.6633$) sets for the combination of audio and video.

\subsection{\ac{MuSe-Reaction}}
\Cref{tab:reaction} shows the results for the \ac{MuSe-Reaction} baseline.
As expected, the audio results are substantially lower than those from the video modality. Of particular note, as it pertains to audio, we see that the emotion-tailored feature-set of \ac{eGeMAPS} performs poorly, almost $0.05$ $\rho$ lower on the development set  than the \ds{} features. Given that there is limited speech in the data set, this may be why the \ds{} features perform better, as due to being spectrogram-based, they can potentially capture a more general acoustic scene and non-speech verbalisations potentially better. %

For the video features, the \ac{FAU}s are performing much better on the test set than \vggf{} (although both are derived from faces), given the nature of the data being `reactions', it may be that the facial action units are much more dynamic generally, and these features model more accurately the emotional expression occurring within the scene.

\begin{figure*}[h!]
    \centering
    \subfloat[Amusement (\ac{FAU}  $\rho$ .405)]{\includegraphics[width=0.49\linewidth]{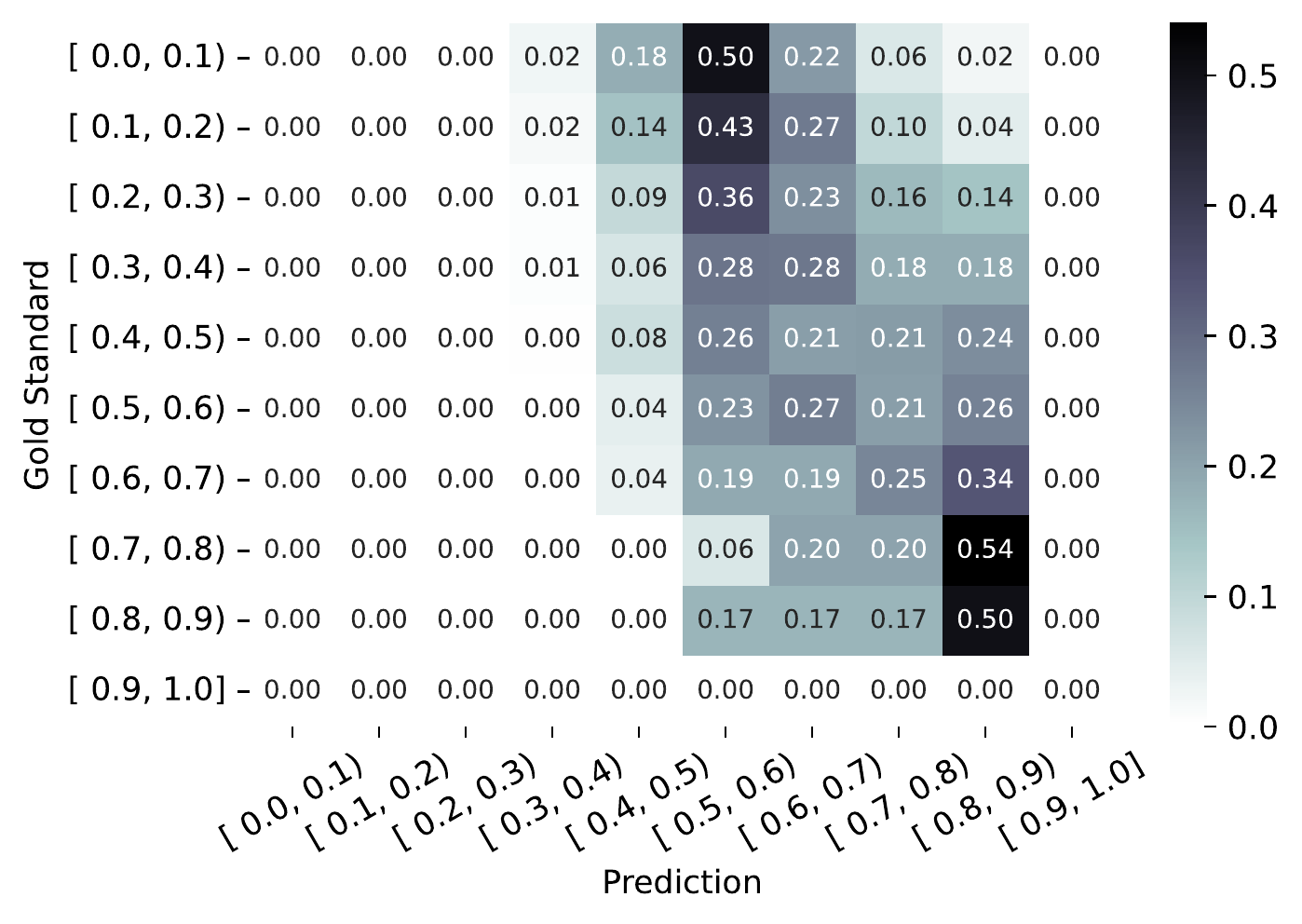}}
    \label{cm-amusement}
    \subfloat[Disgust (\ac{FAU}  $\rho$ .171 )]{\includegraphics[width=0.49\linewidth]{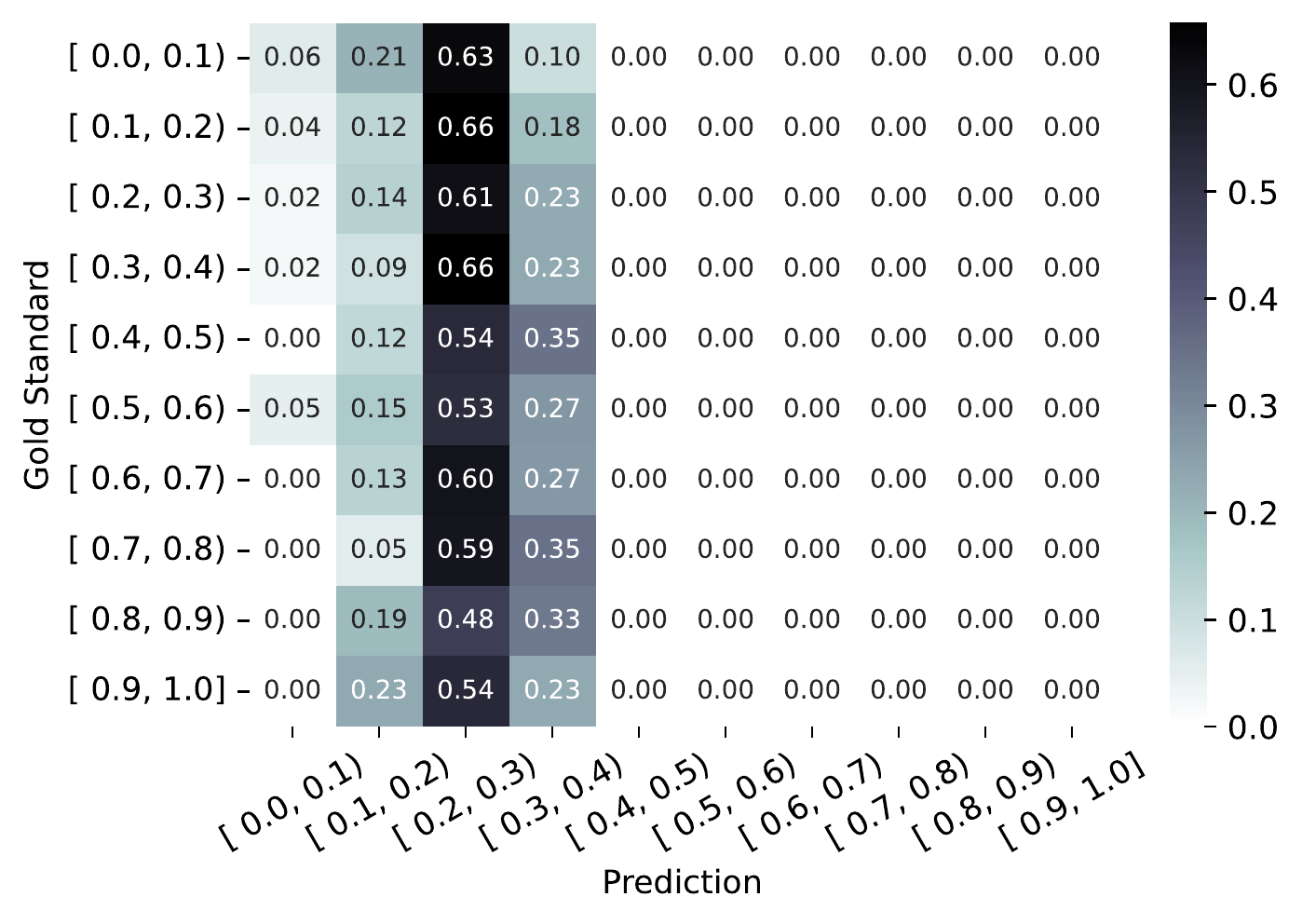}}
    \label{cm-disgust}
    \caption{Confusion matrices for the best (Amusement) and worst (Disgust) performing classes for the best test set configurations in \ac{MuSe-Reaction} as reported in~\Cref{tab:reaction}.}
    \label{fig:cms_reactions}
\end{figure*}

Interestingly, when we observe the individual class scores, we see that \textit{Amusement} is consistently performing better than all other classes, a finding which is consistent for audio and video features (\ac{eGeMAPS}: .148 $\rho$, and \ac{FAU}: .405 $\rho$). As well as being the most likely class to contain non-verbal communication \eg laughter, this performance may be due to the known ease of modelling highly aroused states of emotional expression~\cite{tzirakis2018end2you}. However, it may also relate to the valence of the emotions as we can see from \Cref{fig:cms_reactions}, the \textit{Disgust} class is the worst performing for \ac{FAU}. 

It is worth noting that in this case, the early-fusion of the two best-performing feature sets in each modality does not yield any beneficial results. This holds, although we do consider that through the use of a knowledge-based audio approach, we may see more improvement for audio, which may result in stronger performance via fusion.

\begin{table}[hbt!]
\caption{Results for \ac{MuSe-Reaction}. Reported is the mean Pearson's Correlation Coefficient ($\rho$) for the 7 emotional reaction classes. For each feature and late
fusion configuration, the result for the best of 5 fixed seeds is given. The respective mean and standard deviation of the
results are provided in parentheses.}

\resizebox{1\columnwidth}{!}{%
 \begin{tabular}{lcc}
 \toprule 
 & \multicolumn{2}{c}{[$\rho$]} \\
 Features & \multicolumn{1}{c}{Development} & \multicolumn{1}{c}{Test}  \\ \midrule \midrule
 \multicolumn{3}{l}{\textbf{Audio}} \\
 \ac{eGeMAPS} & .0583 (.0504 $\pm$ .0069) & .0552 (.0479 $\pm$ .0062) \\
 \ds & .1087 (.0945 $\pm$ .0096) & .0741 (.0663 $\pm$ .0077) \\
 \midrule
  \multicolumn{3}{l}{\textbf{Video}} \\
 \ac{FAU}{} & .2840	 (.2828 $\pm$ .0016) & \textbf{.2801} (.2777 $\pm$ .0017) \\
 \vggf & .2488 (.2441 $\pm$ .0027) & .1830 (.1985 $\pm$ .0088) \\
 \midrule
 \multicolumn{3}{l}{\textbf{Early Fusion}} \\
 A+V & .2382	 (.2350 $\pm$ 0.0016) & .2029 (.2014 $\pm$ .0086) \\
 
 \bottomrule
 \end{tabular}\label{tab:reaction}
}

\end{table}

\subsection{\ac{MuSe-Stress}}
\Cref{tab:stress} reports the results obtained for \ac{MuSe-Stress}. Consistent with results reported by some of last year’s participants~\cite{hamieh2021multi, duong2021multi}, the results for \ac{MuSe-Stress} partly fail to generalise to the test data. With respect to single modality experiments, this observation is particularly significant for the video features. For example, the best seed for predicting physiological arousal based on Facial Action Units yields a \ac{CCC} of $.5191$ on the development set, but only results in a \ac{CCC} of $.0785$ on the test partition. The audio feature sets, in comparison, achieve better generalisation with the most extreme difference between development and test \ac{CCC} being about $.12$ (for \ac{eGeMAPS} on physiological arousal). Moreover, for both prediction targets, the \ds{}
audio features perform best among the unimodal approaches with
\ac{CCC} values of $.4239$ and $.4931$ on the test sets for physiological
arousal and valence, respectively.

A surprising aspect of the unimodal results is that audio features yield better results for valence than for arousal, contrary to previous results in the domain of multimodal emotion recognition. For the visual features, no such tendency to work better for one of the two dimensions can be observed: \acp{FAU} lead to better results for predicting valence (mean \ac{CCC} of $.3878$ on the test set) than for physiological arousal ($.1135$); the same is true for the \vggf{} features ($.1968$ and $.1576$ mean \ac{CCC} on the test set for valence and arousal, respectively). The textual \bert features account for higher \acp{CCC} on the development partition for valence (mean \ac{CCC} of $.3221$) than for physiological arousal ($.2828$). 
Surprisingly, however, for arousal, they generalise better to the test data, while for valence, the mean \bert \acp{CCC} drops from $.3221$ to $.1872$ when evaluating on the test set.
These partly counterintuitive results may be attributed to the job interview setting. Job interviewees typically suppress nervousness in an attempt to give a relaxed, sovereign impression. This might make the detection of arousal from audio and video difficult. The comparably stable performance of textual features for physiological arousal may be due to correlations between participants pausing their speech for a longer time -- or hardly at all -- and arousal. We find such correlations to exist for several participants.

We also experiment with the downsampled biosignals, motivated by some of last year's approaches (\cite{ma2021hybrid,zhang2021multimodal,cai2021multimodal}) to the task which used these signals as a feature. To do so, we concatenate the three signals (BPM, ECG, and respiratory rate) into a three-dimensional feature vector and normalise them. Here, severe generalisation and stability problems can be observed. To give an example, for arousal, the mean \ac{CCC} performance of biosignal features on the development set is $.2793$, but for the test set, it drops to .1095. What is more, the standard deviations obtained with the biosignal results are consistently higher than those of any other modality. Because of these issues and in order not to inflate the number of experiments, we exclude the physiological modality from the late fusion experiments.

While valence prediction could not be improved by late fusion, the late fusion of the audio and text modality accounts for the best result on the test set for physiological arousal prediction ($.4761$ CCC), slightly surpassing the late fusion of audio and text ($.4413$) as well as \ds{} ($.4239$). For valence, a generalisation issue for late fusion is apparent. To give an example, the late fusion of acoustic and visual features yields by far the best result on the development set ($.6914$) but only achieves a \ac{CCC} of $.4906$ on the test set. %
\begin{table*}[h!bt]
\caption{Results for \ac{MuSe-Stress}. Reported are the \ac{CCC} values for valence, and physiological arousal. For each feature and late fusion configuration, the result for the best of 20 fixed seeds is given. The respective mean and standard deviation of the results are provided in parentheses.
The combined results are the mean of arousal and valence test \acp{CCC} for each feature set. 
}

\resizebox{1.0\linewidth}{!}{%
 \begin{tabular}{lccccc}
 \toprule
  & \multicolumn{2}{c}{\textbf{(Physiological) Arousal}} & \multicolumn{2}{c}{\textbf{Valence}} & \textbf{Combined} \\ 
  & \multicolumn{2}{c}{[\ac{CCC}]} & \multicolumn{2}{c}{[\ac{CCC}]} & [\ac{CCC}] \\
 Features   & Development & Test   & Development & Test & Test \\ 
 \midrule \midrule

 \multicolumn{6}{l}{\textbf{Audio}} \\
  \ac{eGeMAPS}   & .4112 (.3168 $\pm$ .0459) & .2975 (.3338 $\pm$ .0836) & .5090 (.4744 $\pm$ .0244) & .3988 (.3932 $\pm$ .0385) & .3482 \\
  \ds   & .4139 (.3433 $\pm$ .0548) & .4239 (.4372 $\pm$ .0323) & .5741 (.5395 $\pm$ .0207) & \textbf{.4931} (.4826 $\pm$ .0324) & \textbf{.4585} \\ 
  \midrule

  \multicolumn{6}{l}{\textbf{Video}} \\
  \ac{FAU}   & .5191 (.4257 $\pm$ .0475) & .0785 (.1135 $\pm$ .0335) & .4751 (.3886 $\pm$ .0534) & .2388 (.3878 $\pm$ .0560) & .1918 \\
  \vggf   & .3171 (.2697 $\pm$ .0216) & .2076 (.1576 $\pm$ .0285) & .2637 (.1106 $\pm$ .0739) & .0936 (.1968 $\pm$ .1130) & .1506 \\
    \midrule

  \multicolumn{6}{l}{\textbf{Text}} \\
  \bert   & .3280 (.2828 $\pm$ .0372) & .3504 (.3218 $\pm$ .0423) & .3672 (.3221 $\pm$ .0285) & .1864 (.1872 $\pm$ .0269) & .2683 \\
    \midrule

  \multicolumn{6}{l}{\textbf{Physiological}} \\
  BPM + ECG + resp.   & .3917 (.2793 $\pm$ .0782) & .1095 (.1151 $\pm$ .0656) & .4361 (.2906 $\pm$ .0787) & .1861 (.2141 $\pm$ .0953) & .1478 \\
    \midrule

  \multicolumn{6}{l}{\textbf{Late Fusion}} \\
 A+T   & .4478 (.4409 $\pm$ .0038) & \textbf{.4761} (.4716 $\pm$ .0034) & .5243 (.4808 $\pm$ .0161) & .3653 (.3163 $\pm$ .0211) & .4207 \\
  A+V   & .5440 (.5167 $\pm$ .0142) & .3777 (.4011 $\pm$ .0229) & .6914 (.6811 $\pm$ .0081) & .4906 (.4969 $\pm$ .0184) & .4342 \\
  T+V   & .4609 (.4425 $\pm$ .0112) & .3303 (.3327 $\pm$ .0112) & .5144 (.4965 $\pm$ .0102) & .2462 (.2364 $\pm$ .0082) & .2883 \\
  A+T+V   & .5056 (.4940 $\pm$ .0070) & .4413 (.4485 $\pm$ .0125) & .6104 (.5720 $\pm$ .0215) & .3703 (.3455 $\pm$ .0258) & .4058 \\

 \bottomrule 
 \end{tabular}
}
\label{tab:stress}
\end{table*}

\section{Conclusions}\label{sec:conclusion}
This baseline paper introduced MuSe 2022 -- the 3rd Multimodal Sentiment Analysis challenge. MuSe 2022 features three multimodal datasets: \ac{Passau-SFCH} with press conference recordings of football coaches annotated for humour, \ac{MuSe-Reaction} containing emotional reactions to stimuli, and \ac{Ulm-TSST} consisting of recordings of the stress-inducing TSST. The challenge offers three sub-challenges accounting for a wide range of different prediction targets: i) in \ac{MuSe-Humor}, humour in press conferences is to be detected; ii) in \ac{MuSe-Reaction}, 
the intensities of 7 emotion classes are to be predicted; 
and iii) \ac{MuSe-Stress} is a regression task on the levels of continuous valence and arousal values in a stressful situation. Similar to previous iterations (\cite{stappen2020muse1, stappen2021muse}), we employed open-source software to provide participants with an array of extracted features in order to facilitate fast development of novel methods. Based on these features, we set transparent and realistic baseline results. Features, code, and raw data are made publicly available. The official baselines on the test sets are as follows: $.8480$ \ac{AUC} for \ac{MuSe-Humor} as achieved using \vggf{} features; a mean $\rho$ over all classes of $.2801$ for \ac{MuSe-Reaction} is obtained utilising \ac{FAU}, and a \acp{CCC} of $.4761$ and $.4931$ for physiological arousal and valence, respectively, for \ac{MuSe-Stress}, based on \ds{} features and a late fusion of audio and text modalities, respectively.

The provided baselines give a first impression on which features and modalities may be suited best for the different tasks. %
We believe that more refined methods of combining different modalities and features may lead to significant improvements over the reported baseline results. We hope that MuSe 2022 serves as a stimulating environment for developing and evaluating such novel approaches.

\section{Acknowledgments}
This project has received funding from the 
Deutsche Forschungsgemeinschaft (DFG) under grant agreement No.\ 461420398,
and the DFG's Reinhart Koselleck project No.\ 442218748 (AUDI0NOMOUS).

\begin{acronym}
\acro{AReLU}[AReLU]{Attention-based Rectified Linear Unit}
\acro{AUC}[AUC]{Area Under the Curve}
\acro{CCC}[CCC]{Concordance Correlation Coefficient}
\acro{CNN}[CNN]{Convolutional Neural Network}
\acrodefplural{CNN}[CNNs]{Convolutional Neural Networks}
\acro{CI}[CI]{Confidence Interval}
\acrodefplural{CI}[CIs]{Confidence Intervals}
\acro{CCS}[CCS]{COVID-19 Cough}
\acro{CSS}[CSS]{COVID-19 Speech}
\acro{CTW}[CTW]{Canonical Time Warping}
\acro{ComParE}[ComParE]{Computational Paralinguistics Challenge}
\acrodefplural{ComParE}[ComParE]{Computational Paralinguistics Challenges}
\acro{DNN}[DNN]{Deep Neural Network}
\acrodefplural{DNNs}[DNNs]{Deep Neural Networks}
\acro{DEMoS}[DEMoS]{Database of Elicited Mood in Speech}
\acro{eGeMAPS}[\textsc{eGeMAPS}]{extended Geneva Minimalistic Acoustic Parameter Set}
\acro{EULA}[EULA]{End User License Agreement}
\acro{EWE}[EWE]{Evaluator Weighted Estimator}
\acro{FLOP}[FLOP]{Floating Point Operation}
\acrodefplural{FLOP}[FLOPs]{Floating Point Operations}
\acro{FAU}[FAU]{Facial Action Unit}
\acrodefplural{FAU}[FAUs]{Facial Action Units}
\acro{GDPR}[GDPR]{General Data Protection Regulation}
\acro{HDF}[HDF]{Hierarchical Data Format}
\acro{Hume-Reaction}[\textsc{Hume-Reaction}]{Hume-Reaction}
\acro{HSQ}[HSQ]{Humor Style Questionnaire}
\acro{IEMOCAP}[IEMOCAP]{Interactive Emotional Dyadic Motion Capture}
\acro{KSS}[KSS]{Karolinska Sleepiness Scale}
\acro{LIME}[LIME]{Local Interpretable Model-agnostic Explanations}
\acro{LLD}[LLD]{Low-Level Descriptor}
\acrodefplural{LLD}[LLDs]{Low-Level Descriptors}
\acro{LSTM}[LSTM]{Long Short-Term Memory}
\acro{MIP}[MIP]{Mood Induction Procedure}
\acro{MIP}[MIPs]{Mood Induction Procedures}
\acro{MLP}[MLP]{Multilayer Perceptron}
\acrodefplural{MLP}[MLPs]{Multilayer Perceptrons}
\acro{MPSSC}[MPSSC]{Munich-Passau Snore Sound Corpus}
\acro{MTCNN}[MTCNN]{Multi-task Cascaded Convolutional Networks}
\acro{MuSe}[MuSe]{\textbf{Mu}ltimodal \textbf{Se}ntiment Analysis Challenge}
\acro{MuSe-Humor}[\textsc{MuSe-Humor}]{Humor Detection Sub-Challenge}
\acro{MuSe-Reaction}[\textsc{MuSe-Reaction}]{Emotional Reactions Sub-Challenge}
\acro{MuSe-Stress}[\textsc{MuSe-Stress}]{Emotional Stress Sub-Challenge}
\acro{Passau-SFCH}[\textsc{Passau-SFCH}]{Passau Spontaneous Football Coach Humor}
\acro{RAAW}[\textsc{RAAW}]{Rater Aligned Annotation Weighting}
\acro{RAVDESS}[RAVDESS]{Ryerson Audio-Visual Database of Emotional Speech and Song}
\acro{SER}[SER]{Speech Emotion Recognition}
\acro{SHAP}[SHAP]{SHapley Additive exPlanations}
\acro{SLEEP}[SLEEP]{Düsseldorf Sleepy Language Corpus}
\acro{STFT}[STFT]{Short-Time Fourier Transform}
\acrodefplural{STFT}[STFTs]{Short-Time Fourier Transforms}
\acro{SVM}[SVM]{Support Vector Machine}
\acro{TF}[TF]{TensorFlow}
\acro{TSST}[TSST]{Trier Social Stress Test}
\acro{TNR}[TNR]{True Negative Rate}
\acro{TPR}[TPR]{True Positive Rate}
\acro{UAR}[UAR]{Unweighted Average Recall}
\acro{Ulm-TSST}[\textsc{Ulm-TSST}]{Ulm-Trier Social Stress Test}
\acrodefplural{UAR}[UARs]{Unweighted Average Recall}
\end{acronym}

\clearpage
\footnotesize
\bibliographystyle{ACM-Reference-Format}
\balance
\bibliography{sample-base}

\end{document}